\documentclass[11pt,a4paper]{article}
\usepackage[hyperref]{AACL-IJCNLP2020}
\usepackage{times}
\usepackage{latexsym}

\usepackage{microtype}
\usepackage{listings}
\usepackage[utf8]{inputenc}
\usepackage[vietnamese,english]{babel}
\usepackage{CJKutf8}
\usepackage[T1]{fontenc}
\usepackage{multirow}
\usepackage{color}
\usepackage{listings}
\usepackage{amsmath,amssymb}

\definecolor{codegreen}{rgb}{0,0.6,0}
\definecolor{codegray}{rgb}{0.5,0.5,0.5}
\definecolor{codepurple}{rgb}{0.58,0,0.82}
\definecolor{backcolour}{rgb}{0.95,0.95,0.92}

\lstdefinestyle{mystyle}{
    backgroundcolor=\color{backcolour},   
    commentstyle=\color{codegreen},
    keywordstyle=\color{magenta},
    numberstyle=\tiny\color{codegray},
    stringstyle=\color{codepurple},
    basicstyle=\footnotesize,
    breakatwhitespace=false,         
    breaklines=true,                 
    captionpos=b,                    
    keepspaces=true,                            
    showspaces=false,                
    showstringspaces=false,
    showtabs=false,                  
    tabsize=2
}
\aclfinalcopy 
\def\aclpaperid{13X-D9J6G6C9J6} 

\newcommand{\vect}[1]{\mathbf{#1}}
\title{Improving Multilingual Neural Machine Translation For Low-Resource Languages: French, English - Vietnamese}

\author{
    Thi-Vinh Ngo \\
  Thai Nguyen University \\
  \texttt{ntvinh@ictu.edu.vn} \\  \And
    Phuong-Thai Nguyen \\
   Vietnam National University \\
  \texttt{npthai@vnu.edu.vn } \\
  \AND   
  Thanh-Le Ha \\
  Karlsruhe Institute of Technology \\
  \texttt{thanh-le.ha@kit.edu} \\  
  \And 
  \hspace{0.5cm } Khac-Quy Dinh \\
  \hspace{0.5cm } Vietnam National University \\
  \hspace{0.3cm } \texttt{moduledk@gmail.com} \\ \And
  Le-Minh Nguyen  \\
  JAIST, Japan  \\
  \texttt{nguyenml@jaist.ac.jp}
  \\}

\date{}

\begin{document}
\maketitle
\begin{abstract}

Prior works have demonstrated that a low-resource language pair can benefit from multilingual machine translation (MT) systems, which rely on many language pairs' joint training. This paper proposes two simple strategies to address the rare word issue in multilingual MT systems for two low-resource language pairs: French-Vietnamese and English-Vietnamese. The first strategy is about dynamical learning word similarity of tokens in the shared space among source languages while another one attempts to augment the translation ability of rare words through updating their embeddings during the training. Besides, we leverage monolingual data for multilingual MT systems to increase the amount of synthetic parallel corpora while dealing with the data sparsity problem. We have shown significant improvements of up to +1.62 and +2.54 BLEU points over the bilingual baseline systems for both language pairs and released our datasets for the research community.
\end{abstract}

\section{Introduction}

Neural Machine Translation (NMT) \cite{Bahdanau2015} has achieved state of the art in various MT systems, including rich and low resource language pairs \cite{Edunov2018, Gu2019, ngo2019}.
However, the quality of low-resource MT is quite unpretentious due to the lack of parallel data while it has achieved better results on systems of the available resource. Therefore, low-resource MT is one of the essential tasks investigated by many previous works \cite{ Ha2016, Lee2016, senrich2019}.

Recently, some works present MT systems that have achieved remarkable results for low-resource language \cite{Gu2019, Roee2020}. Inspired by these works, we collect data from the TED Talks domain, then attempt to build multilingual MT systems from French, English-Vietnamese. Experiments demonstrate that both language pairs: French-Vietnamese and English-Vietnamese have achieved significant performance when joining the training.

Although multilingual MT can reduce the sparse data in the shared space by using word segmentation, however, rare words still exist, evenly they are increased more if languages have a significant disparity in term vocabulary. Previous works suggested some strategies to reduce rare words such as using translation units at sub-word and character levels or generating a universal representation at the word and sentence levels \cite{ Lee2016, Gu2019}. These help to downgrade the dissimilarity of tokens shared from various languages. However, these works require learning additional parameters in training, thus increasing the size of models.

Our paper presents two methods to augment the translation of rare words in the source space without modifying the architecture and model size of MT systems: (1) exploiting word similarity. This technique has been mentioned by previous works \cite{luong2015, Li2016, trieu, ngo2019}. They employ monolingual data or require supervised resources like a bilingual dictionary or WordNet, while we leverage relation from the multilingual space of MT systems. (2) Adding a scalar value to the rare word embedding in order to facilitate its translation in the training process. 

Due to the fact that NMT tends to have bias in translating frequent words, so rare words (which have low frequency) often have less opportunity to be considered. Our ideal is inspired by the works of \cite{Toan2017, ngo2019, Gu2019}. \cite{Toan2017} and \cite{ngo2019} proposed various solutions to urge for translation of rare words, including modification embedding in training. They only experimented with recurrent neural networks (RNNs) while our work uses the state-of-the-art transformer architecture. \cite{Gu2019} transforms the word embedding of a token into the universal space, and they learn plus parameters while our method does not.  We apply our strategies in our fine-tuning processes, and we show substantial improvements of the systems after some epochs only.

Monolingual data are widely used in NMT to augment data for low-resource NMT systems \cite{Sennrich2015, zhang2016,lample2018unsupervised, wu2019, siddhant2020leveraging}. Back-translation \cite{Sennrich2015} is known as the most popular technique in exploiting target-side monolingual data to enhance the translation systems while the self-learning method \cite{zhang2016} focuses on utilizing source-side monolingual data. Otherwise, the dual-learning strategy \cite{wu2019} also suggests using both source- and target-side monolingual data to tackle this problem. Our work investigates the self-learning method \cite{zhang2016} on the low-resource multilingual NMT systems specifically related to Vietnamese. Besides, monolingual data are also leveraged in unsupervised\cite{lample2018unsupervised} or zero-shot translation\cite{lample2018unsupervised}.


The main contributions of our work are:
\vspace*{-0.2cm}
\begin{itemize}
\setlength{\itemsep}{0pt}
  \item We first attempt to build a multilingual system for two low-resource language pairs: French-Vietnamese and English-Vietnamese.
  \item We propose two simple techniques to encourage the translation of rare words in multilingual MT  to upgrade the systems. 
  \item We investigate the quality translation of the low-resource multilingual NMT systems when they are reinforced synthetic data.
  \item We release more datasets extracted from the TED Talks domain for the research purpose: French-Vietnamese and English-Vietnamese.
\end{itemize}

In section 2, we review the transformer architecture used for our experiments. The brief of multilingual translation is shown in section 3. Section 4 presents our methods to deal with rare words in multilingual translation scenarios. The exploitation of monolingual data for low-resource multilingual MT is discussed in section 5. Our results are described in section 6, and related work is shown in section 7. Finally, the paper ends with conclusions and future work.
\section{Transformer-based NMT}
Transformer architecture for machine translation is mentioned for the first time by \cite{Vaswani2017}. This is based on the sequence to sequence framework \cite{Sutskever2014} which includes an encoder to transform information of the source sentence $X=(x_1, x_2,...,x_n)$ into continuous representation and a decoder to generate  the target sentence $Y=(y_1,y_2,...,y_m)$.

Self-attention is an important mechanism in the transformer architecture. It enables the ability to specify the relevance of a word with the remaining words in the sentence through the equation:
\begin{equation}
\begin{aligned}
\text{Self-Attn}(\vect{Q},\vect{K},\vect{V}) =\text {Softmax} (\displaystyle \frac{\vect{Q}\vect{K}^T}{d}) \vect{V}
\end{aligned}
\label{eq:att}
\end{equation}
where \textit{K} (key),\textit{ Q} (query), \textit{V }(value) are the representations of the input sentence and  \textit{d} is the size of the input. 
The attention mechanism \cite{Luong2015a} bridges between the source sentence in the encoder and the target sentence in the decoder. Furthermore, the feed-forward networks are used to normalize the outputs on both encoder and decoder.

The MT system is trained to minimize the maximum likelihood of \textit{K} parallel pairs:
\begin{equation}
\begin{aligned}
\mathcal{L} (\theta)= \frac{1}{K} \sum_{k=1}^{k=K} logp(Y^k|X^k; \theta)
\end{aligned}
\label{lagra}
\end{equation}

\section{Multilingual NMT}
Multilingual NMT systems can translate between many language pairs, even in the zero-shot issue. Previous works investigate multilingual translation in many fashions: (1) Many to many  \cite{Ha2016, Roee2020}: from many sources to many target languages; (2) Many to one \cite{Gu2019}: from many source languages to a target language; (3) One to many \cite{Wang2018}: from one source language to many target languages.  In cases (1) and (3), an artificial token is often added to the beginning of the source sentence to specify the predicted target language. Our MT systems are the same as the case (2), so we do not add any artificial token to the texts.

In a multilingual NMT system from many to one with $M$ language pairs and $K$ sentence pairs for each one, the objective function uses maximum likelihood estimation on the whole parallel pairs $\left\{ X^{(m,k)}, Y^{(m,k)}\right\} _{k=1..K}^{m=1...M} $ as:
\begin{equation}
\begin{aligned}
\mathcal{L} (\theta)= \frac{1}{K} \sum_{m=1}^{m=M} \sum_{k=1}^{k=K} logp(Y^{(m,k)}|X^{(m,k)}; \theta)
\end{aligned}
\label{lagra}
\end{equation}
where $K=\sum_{m=1}^{m=M} K_m$ is the total number of sentences of the whole corpus.

The vocabulary of the source side is mixed from all source languages: $V=\sum_{m=1}^{m=M} V_m$.

\cite{Gu2019} has shown that if the languages shared the same alphabet and had many similar words, such system will get many advantages from multilingual MT. In fact, different words from many languages can differ in form, but they may share the same subwords. This significantly reduces the number of rare words in the MT systems. Nevertheless, the rare word issue is still a challenge in NMT. We choose  English and French are source languages in our experiment with the hope that they can share many tokens even though we do not have much data of those translation directions. 

\section{Augmenting Rare Word Translation} 
\label{methods}
\subsection{Learning multilingual word similarity}
\label{ws}
We assume that a rare word or rare token (which has a low frequency in the training data) from one source language may be similar to another word in a shared multilingual space. Similar words can belong to several languages and they can be replaced by the others.

Our method replaces rare tokens with their similar tokens in shared space. The replacements are learned dynamically in the training NMT system. To avoid slowing down the training speed, we only compute similar tokens after each epoch. In the experiments, we attempt to replace rare tokens from French with similar tokens in English and French.

Our method is described as follows: 

Firstly, we extract the lists of all tokens from the English - $\{A\}$ corpus, and the most \textit{k} common words  from the vocabulary of the source side of the French - $\{B\}$.  We set \textit{k=15} thousand words in the experiments.

Secondly, we compute the similarity score between the embedding of a rare token $ t_i $, $ \forall t_i \notin \{ A \cup B \} $ and each embedding of the tokens $ t_j $, $ \forall t_j \in \{ A \cup B \} $ as follows:
\begin{equation}
\begin{aligned}
  score_i= min(d_j(\vect{e_i},\vect{e_j}) \cdot e^{\cos(\vect{e_i}, \vect{e_j})}) 
\end{aligned}
\label{score}
\end{equation}

where $j=1..M$ with $M$ is the number of tokens of $ {A \cup B} $;  \textit{d} is the Euclidean distance between embedding $e_i$ of token $t_i$ and embedding $e_j$ of token $t_j$.

The last, the token $ t_i $ is replaced by its similar tokens. The scores are computed iteratively after each epoch during the training process. It may have more tokens similar to a rare token, so we experimentalize in the case of random selection a token from the similar tokens. To accrete the effectiveness of the method, we use a threshold to neglect similar pairs that have scores close to 0 or too large. In the experiments, we choose the scores in $[2.4, 2.72]$ to warrant similar pairs alike in terms of distance as well as direction.

\subsection{Updating source embedding}
\label{se}

In this approach, we assume that the embedding $ e_i $ of token $t_i$, $ \forall t_i \notin \{ A \cup B \} $ is represented by the approximate embedding vector as following: 

\begin{equation}
\begin{aligned}
  \vect{e_i}= \vect{e_i} + d  \\
\end{aligned}
\label{score}
\end{equation}
where $d$ is the difference between embedding $ e_i $ and the average of the all embeddings $ e_j $ of token $t_j$, $ \forall t_j \in \{ A \cup B \} $:
\begin{equation}
\begin{aligned}
  d = \vect{e_i} - \frac{\sum_{j=1}^{j=M} \vect{e_j}}{M} 
\end{aligned}
\label{score}
\end{equation}
where M is the number of tokens of $ \{A \cup B \} $.

These embeddings are then updated during the training. The average of embeddings is only estimated after each epoch to avoid slowing down the training speed. We observe the improvements in both language pairs in the experiments. 

\section{Exploiting monolingual data for low-resource multilingual NMT}
\label{monolingual} 

Similar to the idea suggested in \cite{zhang2016}, we leverage monolingual data from the source-side to generate synthetic bilingual data. Instead of using monolingual data from all source languages, we only attempt to exploit monolingual data of English. 

Firstly, we train the multilingual NMT system from English, French $ \rightarrow $ Vietnamese based on bilingual data from the TED talks with the approaches mentioned in section \ref{methods}. The best system is then used to translate English to Vietnamese.

Lastly, the synthetic parallel data are mixed with original bilingual data in the normal training scheme.

\section{Experiments}

\subsection{Datasets}

We extracted data from TED Talks domain\footnote{\url{https://www.ted.com/}} for two language pairs English-Vietnamese and French-Vietnamese. The details of those datasets are described in Table \ref{tab1}. For the English-Vietnamese, we used standard datasets like {\tt tst2012} and {\tt tst2013} from \cite{cettolo2016iwslt} as dev and test sets for validation and evaluation. For French-Vietnamese, we separate a subset from collected data for the same purposes.

\vspace{0.4cm}
\begin{table} [h]
\vspace*{-0.1cm}
\centerline{
\begin{tabular}{|c|c|c|c|}
\hline 
Datasets & Training  &  dev & test \\
\hline
 English-Vietnamese & 231K & 1553 & 1268 \\
 \hline
 French-Vietnamese & 203K & 1007 & 1049 \\
 \hline
\end{tabular}}
\caption{\label{tab1} {The bilingual datasets in our experiments}}
\vspace*{-0.4cm}
\end{table}
\vspace{0.2cm}

To generate synthetic bilingual data, we sampled 1.2 millions English monolingual sentences from the European Parliament English-French corpus\footnote{\url{https://www.statmt.org/europarl}}.  After inferring from the multilingual MT system, we obtained two sets of pseudo bilingual data: English - Vietnamese, French - Vietnamese.

\subsection{Preprocessing} 

English and French texts were tokenized and true-cased using Moses's scripts, and then they are applied to Sennrich's BPE \cite{Sennrich2016}. 30000 operators are learned to generate BPE codes for both languages.

For Vietnamese texts, we only did tokenization and true-casing using Moses's scripts.

We extracted a list of all tokens in English (A) and another list of the 15K most frequency of tokens in French (B). All lists were then used for the mentioned strategies in section~\ref{methods}.

\subsection{Systems and Training} 
We implement our NMT systems using the framework {\tt NMTGMinor}\footnote{\url{https://github.com/quanpn90/NMTGMinor}}. The same settings are used for all experiments. The system includes 4 layers for both encoder and decoder, and the embedding size is 512. For the systems that adapted monolingual data, we use 6 layers. Adam optimizer is set with the initial learning rate at 1.0 for baseline and the multilingual systems and 0.5 for the fine-tuned systems. The size of a mini-batch is 128, and the vocabulary size is set to be the top 50K most frequent tokens. Training and development sets of both language pairs are concatenated prior to the training of our multilingual systems.

We modified this framework to apply our ideals proposed in section \ref{methods}. To speed up the training, we compute the similarity scores and find out similar tokens for rare tokens or the mean of all tokens in $ \{ A \cup B \} $ after each epoch. We replace rare tokens or update their embeddings in each batch. We do not use these techniques for the decoding process, so the system's performance is not affected.

The baseline and multilingual systems are trained for 70 epochs. Our methods are then used to fine-tune the systems for 15 epochs. We choose the five best models to decode the test sets independently for residual systems despite the baseline systems. The beam size is 10, and we try different values of \textit{alpha}: $0.2, 0.4, 0.8, 1.0$. Other settings are the default settings of {\tt NMTGMinor}.

\subsection{Results}

\begin{center}
\begin{table*}[t]
\vspace*{-0.1cm}
{\small
\hfill{}
\begin{tabular}{|c|l|c|c|}
\hline 
\textbf{Datasets} & \textbf{Systems}  & \textbf{ dev} & \textbf{test} \\
\hline
 \multirow{5}{*}{English $\rightarrow$ Vietnamese} & 
  Bilingual Baseline & 31.74 & 35.13  \\
\cline{2-4}
  & Multilingual  & 31.66 (-0.08) &  \textbf{36.18 (+1.05)}\\
 \cline{2-4}
  & Multilingual + fine-tuning & \textbf{31.88 (+0.14)} & \textbf{36.56 (+1.43)} \\
 \cline{2-4}
  & Multilingual + fine-tuning with similarity & \textbf{31.93 (+0.19)} & \textbf{36.75 (+1.62)} \\
 \cline{2-4}
  & Multilingual + fine-tuning with updated embedding & \textbf{32.11 (+0.37)} & \textbf{36.74 (+1.61)}\\
  \cline{2-4}
  & Multilingual + mixing pseudo bilingual data & 30.86 (-0.88) &  35.09 (-0.04) \\ 
 \hline
 \hline
 \multirow{5}{*} { French $\rightarrow$ Vietnamese}  & Bilingual Baseline & 23.07  & 23.03  \\
 \cline{2-4}
  & Multilingual & \textbf{24.49 (+1.42)} & \textbf{24.22 (+1.19)} \\
 \cline{2-4}
  & Multilingual + fine-tuning & \textbf{24.51 (+1.44)}  & \textbf{24.86 (+1.83)} \\
 \cline{2-4}
  & Multilingual + fine-tuning with similarity & \textbf{24.37 (+1.30)} & \textbf{24.70 (+1.63)}
 \\
 \cline{2-4}
 & Multilingual + fine-tuning with updated embedding & \textbf{24.60 (+1.53)}  & \textbf{24.96 (+1.93)}
 \\
 \cline{2-4}
  & Multilingual + mixing pseudo bilingual data & \textbf{25.59 (+2.52)} &  \textbf{25.57 (+2.54)} \\ 
  \cline{2-4}
  & Pseudo bilingual data translation & 19.00 & 18.71  \\ 
 \hline
 
\end{tabular}}
\hfill{}
\caption{\label{tab2} {The results of our MT systems are measured in BLEU. We evaluate the best model for the baseline systems and the average scores on the five best models for the multilingual and pseudo systems.}}

\end{table*}
\end{center}

We evaluate the quality of systems on two translation tasks: French to Vietnamese and English to Vietnamese, using on different approaches mentioned in previous sections. The {\tt multi-BLEU} from Moses's scripts\footnote{\url{https://github.com/moses-smt/mosesdecoder/tree/master/scripts}} is used. The results have shown in the Table~\ref{tab2}. 

\textbf{(1) Bilingual baseline systems.} We train the systems based on separate bilingual data of each language pair for 70 epochs. The best model is used to decode the test data for comparison purposes in our experiments.

\textbf{(2) Multilingual systems.} We concatenate training and development sets in order to construct the new sets: French, English $\rightarrow$ Vietnamese, and then train the system using those data for the same number of epochs as for the baseline systems. We observe an improvement of  +1.05 BLEU  points on English $\rightarrow$ Vietnamese translation task and another one of +1.19 BLEU  points on French $\rightarrow$ Vietnamese translation task compared to the baseline systems.

\textbf{(3) Multilingual fine-tuning systems.} The multilingual system is fine-tuned from the baseline for further 15 epochs with an initial learning rate of 0.05. We see the improvements of +1.43 and +1.83 BLEU  points on both translation tasks, respectively.

\textbf{(4) Multilingual fine-tuning with similarity systems.} The systems from (2) are fine-tuned with the strategy  mentioned in section \ref{ws} using the modified framework. We obtained a bigger gain of  +1.62 BLEU  points on the English $\rightarrow$ Vietnamese translation task whilst the French $\rightarrow$ Vietnamese translation task has achieved a lower improvement than the systems in (3). We show that the English $\rightarrow$ Vietnamese translation task has more advantages when rare tokens from French are replaced by similar tokens in the multilingual space. In the future, we would attempt the inverse replacement.

\textbf{(5) Multilingual fine-tuning with updated embedding systems.}  We use the modified framework to fine-tune the systems in (2) with the method mentioned in section \ref{se}. The greater improvements can be found at +1.61 and +1.93 on both translation tasks compared to the systems which do not use our methods. 

\textbf{(6) Multilingual with mixing of pseudo bilingual data.} 
We use 400K synthetic bilingual sentence pairs for each of the language pairs: English-Vietnamese and French-Vietnamese. We train the multilingual NMT system on a mix of pseudo and real bilingual data mentioned in section \ref{monolingual} for 50 epochs. And then it is fine-tuned on the actual parallel data for 20 epochs. We observed a bigger improvement of \textbf{+2.54} BLEU  points on the French  $\rightarrow$ Vietnamese system while the English $\rightarrow$ Vietnamese system has achieved less improvement compared to previous systems. We speculate that the English $\rightarrow$ Vietnamese translation task may be affected by the French $\rightarrow$ Vietnamese pseudo bilingual data. In future work, we would leverage the data selection methods in order to equip better synthetic data for our systems.

\textbf{(7) Pseudo bilingual data translation.}
We train the French $\rightarrow$ Vietnamese NMT system relied on only 1.2 thousands pseudo bilingual data mentioned in section \ref{monolingual} for 26 epochs. We achieve 18.71 BLEU points on the averaged model from our five best models. Thus, we can generate synthetic parallel data for a low-resource language pair from another language pair with a bigger bilingual resource.


\section{Related Work} 

Due to the unavailability of the parallel data for low-resource language pairs or zero-shot translation, previous works focus on the task to have more data such as leveraging multilingual translation \cite{Ha2016, ha2017effective, Wang2018, Gu2019, Roee2020} or using monolingual data with back-translation, self-learning \cite{Sennrich2015,zhang2016, wu2019} or mix-source \cite{Ha2016} technique.

For leveraging multilingual translation, \cite{Ha2016} added language code and target forcing in order to learn the shared representations of the source words and specify the target words. \cite{Wang2018} demonstrated a one-to-many multilingual MT with three different strategies which modify their architecture. \cite{Gu2019} built many-to-one multilingual MT systems by adding a layer to transform the source embeddings and representation into a universal space to augment the translation of low resource language, which is similar to ours. \cite{Roee2020} implemented a massive many-to-many multilingual system, employing many low-resource language pairs. All of the mentioned works have shown substantial improvements in low-resource translation, however, they are less correlative to our translation tasks.

Although multilingual MT equips a shared space with many advantages, rare word translation is still the issue that needs to be considered. The task of dealing with rare words has been mentioned in previous works. \cite{luong2015} copied words from source sentences by words from target sentences after the translation using a bilingual dictionary. \cite{Li2016} and \cite{trieu} learned word similarity from monolingual data to improve their systems. Our approach is similar to these works, but we only learn similarity from the shared multilingual space of MT systems. \cite{ngo2019} addressed the rare word problem by using the synonyms from WordNet. 

\cite{Toan2017} and \cite{ngo2019} presented different solutions to solve rare word situation by transforming the embeddings during the training of their RNN-based architecture. Those solutions cannot be applied to the transformer architecture. In \cite{Gu2019}, the embeddings of rare tokens and universal tokens are jointly learned through a plus parameter while we only add a scalar value to the embeddings.

Monolingual data is used to generate synthetic bilingual data in sparsity data issues. \cite{Sennrich2015} proposed back-translation method that uses a backward model to get the source data from the monolingual target data.  In contrast, \cite{zhang2016} shown the self-learning technique by employing a forward model to translate monolingual source data into the target data. \cite{wu2019} incorporated both mentioned techniques into their NMT systems. Monolingual data is also demonstrated its efficiency in unsupervised machine translation\cite{lample2018unsupervised} or in zero-shot multilingual NMT \cite{siddhant2020leveraging, ha2017effective}. In our work, we use the self-learning method to produce pseudo bilingual data, and it is then used to train our low-resource multilingual NMT systems.

\section{Conclusion and Future Work}
We have built multilingual MT systems for two low-resource language pairs: English-Vietnamese and French-Vietnamese, and proposed two approaches to tackle rare word translation. We show that our approaches bring significant improvements to our MT systems. We find that the pseudo bilingual can furthermore enhance a multilingual NMT system in case of French $\rightarrow$ Vietnamese translation task.  In the future, we would like to use more language pairs in our systems and to combine proposed methods in order to evaluate the effectiveness of our MT systems.

\bibliography{aacl-ijcnlp2020}

\begin{thebibliography}{24}
\expandafter\ifx\csname natexlab\endcsname\relax\def\natexlab#1{#1}\fi

\bibitem[{Aharoni et~al.(2019)Aharoni, Johnson, and Firat}]{Roee2020}
Roee Aharoni, Melvin Johnson, and Orhan Firat. 2019.
\newblock \href {http://arxiv.org/abs/1903.00089} {Massively multilingual
  neural machine translation}.
\newblock \emph{CoRR}, abs/1903.00089.

\bibitem[{Bahdanau et~al.(2015)Bahdanau, Cho, and Bengio}]{Bahdanau2015}
Dzmitry Bahdanau, Kyunghyun Cho, and Yoshua Bengio. 2015.
\newblock Neural machine translation by jointly learning to align and
  translate.
\newblock \emph{Proceedings of International Conference on Learning
  Representations}.

\bibitem[{Cettolo et~al.(2016)Cettolo, Niehues, St{\"u}ker, Bentivogli,
  Cattoni, and Federico}]{cettolo2016iwslt}
M~Cettolo, J~Niehues, S~St{\"u}ker, L~Bentivogli, R~Cattoni, and M~Federico.
  2016.
\newblock {The IWSLT 2016 Evaluation Campaign}.
\newblock In \emph{Proceedings of the 13th International Workshop on Spoken
  Language Translation (IWSLT 2016)}, Seattle, WA, USA.

\bibitem[{Edunov et~al.(2018)Edunov, Ott, Auli, and Grangier}]{Edunov2018}
Sergey Edunov, Myle Ott, Michael Auli, and David Grangier. 2018.
\newblock \href {http://arxiv.org/abs/1808.09381} {Understanding
  back-translation at scale}.
\newblock \emph{CoRR}, abs/1808.09381.

\bibitem[{Gu et~al.(2019)Gu, Wang, Cho, and Li}]{Gu2019}
Jiatao Gu, Yong Wang, Kyunghyun Cho, and Victor O.~K. Li. 2019.
\newblock \href {http://arxiv.org/abs/1906.01181} {Improved zero-shot neural
  machine translation via ignoring spurious correlations}.
\newblock \emph{CoRR}, abs/1906.01181.

\bibitem[{Ha et~al.(2017)Ha, Niehues, and Waibel}]{ha2017effective}
Thanh-Le Ha, Jan Niehues, and Alexander Waibel. 2017.
\newblock \href {http://arxiv.org/abs/1711.07893} {{Effective Strategies in
  Zero-Shot Neural Machine Translation}}.

\bibitem[{Ha et~al.(2016)Ha, Niehues, and Waibel}]{Ha2016}
Thanh{-}Le Ha, Jan Niehues, and Alexander~H. Waibel. 2016.
\newblock \href {http://arxiv.org/abs/1611.04798} {Toward multilingual neural
  machine translation with universal encoder and decoder}.
\newblock \emph{CoRR}, abs/1611.04798.

\bibitem[{Lample et~al.(2018)Lample, Conneau, Denoyer, and
  Ranzato}]{lample2018unsupervised}
Guillaume Lample, Alexis Conneau, Ludovic Denoyer, and Marc'Aurelio Ranzato.
  2018.
\newblock \href {http://arxiv.org/abs/1711.00043} {Unsupervised machine
  translation using monolingual corpora only}.

\bibitem[{Lee et~al.(2016)Lee, Cho, and Hofmann}]{Lee2016}
Jason Lee, Kyunghyun Cho, and Thomas Hofmann. 2016.
\newblock \href {http://arxiv.org/abs/1610.03017} {Fully character-level neural
  machine translation without explicit segmentation}.
\newblock \emph{CoRR}, abs/1610.03017.

\bibitem[{Li et~al.(2016)Li, Zhang, and Zong}]{Li2016}
Xiaoqing Li, Jiajun Zhang, and Chengqing Zong. 2016.
\newblock Towards zero unknown word in neural machine translation.
\newblock In \emph{IJCAI}.

\bibitem[{Luong et~al.(2015a)Luong, Pham, and Manning}]{Luong2015a}
Minh{-}Thang Luong, Hieu Pham, and Christopher~D. Manning. 2015a.
\newblock Effective approaches to attention-based neural machine translation.
\newblock In \emph{Proceedings of the 2015 Conference on Empirical Methods in
  Natural Language Processing}, pages 1412--1421.

\bibitem[{Luong et~al.(2015)Luong, Sutskever, Le, Vinyals, and
  Zaremba}]{luong2015}
Thang Luong, Ilya Sutskever, Quoc Le, Oriol Vinyals, and Wojciech Zaremba.
  2015.
\newblock \href {https://doi.org/10.3115/v1/P15-1002} {Addressing the rare word
  problem in neural machine translation}.
\newblock In \emph{Proceedings of the 53rd Annual Meeting of the Association
  for Computational Linguistics and the 7th International Joint Conference on
  Natural Language Processing (Volume 1: Long Papers)}, pages 11--19, Beijing,
  China. Association for Computational Linguistics.

\bibitem[{Ngo et~al.(2019)Ngo, Ha, Nguyen, and Nguyen}]{ngo2019}
Thi-Vinh Ngo, Thanh-Le Ha, Phuong-Thai Nguyen, and Le-Minh Nguyen. 2019.
\newblock \href {https://doi.org/10.18653/v1/D19-5228} {Overcoming the rare
  word problem for low-resource language pairs in neural machine translation}.
\newblock In \emph{Proceedings of the 6th Workshop on Asian Translation}, pages
  207--214, Hong Kong, China. Association for Computational Linguistics.

\bibitem[{Nguyen and Chiang(2017)}]{Toan2017}
Toan~Q. Nguyen and David Chiang. 2017.
\newblock Improving lexical choice in neural machine translation.
\newblock \emph{Proceedings of NAACL-HLT 2018}, pages 334--343.

\bibitem[{Sennrich et~al.(2015)Sennrich, Haddow, and Birch}]{Sennrich2015}
Rico Sennrich, Barry Haddow, and Alexandra Birch. 2015.
\newblock \href {http://arxiv.org/abs/1511.06709} {Improving neural machine
  translation models with monolingual data}.
\newblock \emph{CoRR}, abs/1511.06709.

\bibitem[{Sennrich et~al.(2016)Sennrich, Haddow, and Birch}]{Sennrich2016}
Rico Sennrich, Barry Haddow, and Alexandra Birch. 2016.
\newblock {Neural Machine Translation of Rare Words with Subword Units}.
\newblock In \emph{Association for Computational Linguistics (ACL 2016)}.

\bibitem[{Sennrich and Zhang(2019)}]{senrich2019}
Rico Sennrich and Biao Zhang. 2019.
\newblock \href {http://arxiv.org/abs/1905.11901} {Revisiting low-resource
  neural machine translation: {A} case study}.
\newblock \emph{CoRR}, abs/1905.11901.

\bibitem[{Siddhant et~al.(2020)Siddhant, Bapna, Cao, Firat, Chen, Kudugunta,
  Arivazhagan, and Wu}]{siddhant2020leveraging}
Aditya Siddhant, Ankur Bapna, Yuan Cao, Orhan Firat, Mia Chen, Sneha Kudugunta,
  Naveen Arivazhagan, and Yonghui Wu. 2020.
\newblock \href {http://arxiv.org/abs/2005.04816} {Leveraging monolingual data
  with self-supervision for multilingual neural machine translation}.

\bibitem[{Sutskever et~al.(2014)Sutskever, Vinyals, and Le}]{Sutskever2014}
Ilya Sutskever, Oriol Vinyals, and Quoc~V. Le. 2014.
\newblock \href {http://arxiv.org/abs/1409.3215} {Sequence to sequence learning
  with neural networks}.
\newblock \emph{CoRR}, abs/1409.3215.

\bibitem[{Trieu et~al.(2016)Trieu, Nguyen, and Nguyen}]{trieu}
Hai-Long Trieu, Le-Minh Nguyen, and Phuong-Thai Nguyen. 2016.
\newblock \href {https://www.aclweb.org/anthology/Y16-2024} {Dealing with
  out-of-vocabulary problem in sentence alignment using word similarity}.
\newblock In \emph{Proceedings of the 30th Pacific Asia Conference on Language,
  Information and Computation: Oral Papers}, pages 259--266, Seoul, South
  Korea.

\bibitem[{Vaswani et~al.(2017)Vaswani, Shazeer, Parmar, Uszkoreit, Jones,
  Gomez, Kaiser, and Polosukhin}]{Vaswani2017}
Ashish Vaswani, Noam Shazeer, Niki Parmar, Jakob Uszkoreit, Llion Jones,
  Aidan~N. Gomez, Lukasz Kaiser, and Illia Polosukhin. 2017.
\newblock \href {http://arxiv.org/abs/1706.03762} {Attention is all you need}.
\newblock \emph{CoRR}, abs/1706.03762.

\bibitem[{Wang et~al.(2018)Wang, Zhang, Zhai, Xu, and Zong}]{Wang2018}
Yining Wang, Jiajun Zhang, Feifei Zhai, Jingfang Xu, and Chengqing Zong. 2018.
\newblock \href {https://doi.org/10.18653/v1/D18-1326} {Three strategies to
  improve one-to-many multilingual translation}.
\newblock pages 2955--2960.

\bibitem[{Wu et~al.(2019)Wu, Wang, Xia, Qin, Lai, and Liu}]{wu2019}
Lijun Wu, Yiren Wang, Yingce Xia, Tao Qin, Jianhuang Lai, and Tie-Yan Liu.
  2019.
\newblock \href {https://doi.org/10.18653/v1/D19-1430} {Exploiting monolingual
  data at scale for neural machine translation}.
\newblock In \emph{Proceedings of the 2019 Conference on Empirical Methods in
  Natural Language Processing and the 9th International Joint Conference on
  Natural Language Processing (EMNLP-IJCNLP)}, pages 4207--4216, Hong Kong,
  China. Association for Computational Linguistics.

\bibitem[{Zhang and Zong(2016)}]{zhang2016}
Jiajun Zhang and Chengqing Zong. 2016.
\newblock \href {https://doi.org/10.18653/v1/D16-1160} {Exploiting source-side
  monolingual data in neural machine translation}.
\newblock In \emph{Proceedings of the 2016 Conference on Empirical Methods in
  Natural Language Processing}, pages 1535--1545, Austin, Texas. Association
  for Computational Linguistics.

\end{thebibliography}
\bibliographystyle{acl_natbib}

\end{document}